\documentclass{llncs}
\usepackage[utf8]{inputenc}
\usepackage{graphicx}
\usepackage{tabularx} 
\usepackage{multirow}
\usepackage{cite}
\usepackage{epstopdf}
\usepackage{longtable}
\usepackage{textcomp}

\begin{document}
\title{Graph Centrality Measures for Boosting  Popularity-Based Entity  Linking}
\author{Hussam Hamdan, Jean-Gabriel Ganascia }
\institute{Labex Observatoire de la vie littéraire (OBVIL). Laboratoire d'Informatique de Paris 6 (LIP6), Pierre and Marie Curie University, UMR 7606, 4 place Jussieu, 75005, Paris, France
Hussam.Hamdan@lip6.fr, Jean-Gabriel.Ganascia@lip6.fr}

\maketitle
\begin{abstract}

  Many Entity Linking systems use collective graph-based methods  to disambiguate the entity mentions within a document. Most of them have focused on graph construction and initial weighting of the candidate entities, less attention has been devoted to compare the graph ranking algorithms. In this work, we focus on the graph-based ranking algorithms, therefore we propose to apply five centrality measures: Degree, HITS, PageRank, Betweenness and Closeness. A disambiguation graph of candidate entities is constructed for each document using the popularity method, then  centrality measures are applied to choose the most relevant candidate to boost the results of entity popularity method. We investigate the effectiveness of each centrality measure  on the performance across different domains and datasets. Our experiments show that a simple and fast centrality measure such as Degree centrality can outperform other more time-consuming measures.
\end{abstract}


\section{Introduction}

Named Entity Linking (NEL) task seeks to link the entity mentions in a given text to their relevant entities in a knowledge base (KB).  NEL is  a crucial process   to advance the Web of Linked Data.  It is also useful for many  applications  such as Semantic Search, Text Classification, Reasoning and Question \& Answering. It can enrich the document representation which may improve the performance of many applications.

NEL  is composed of two phases: candidate selection and candidate ranking or disambiguation. A NEL system selects the potential candidates for each named entity mention in a document,  it ranks them according to their relevance in order to link the mention to the most relevant entity. 

Relation between a mention and an entity is many-to-many, each mention denotes many entities and each entity could be denoted by different mentions. For example, the word Paris refers to Paris (capital of France), Paris (a city in USA), Paris (a  prince of Troy in Greek mythology) ...etc, and "The capital of France" can be denoted by Paris, "City of Light" or  "the French Capital" ..etc. NEL task seeks to resolve this ambiguity and assign one entity to one mention according to the context. 

Much  research has parametrised their methods by using supervised approach \cite{hoffart_robust_2011,durrett_joint_2014,ganea_probabilistic_2016,francis-landau_capturing_2016} or unsupervised approach but they  tuned their methods to work only on one dataset \cite{pershina_personalized_2015,alhelbawy_graph_2014}, therefore these methods cannot be considered as a robust solution for different datasets. Other studies have proved the effectiveness of their methods on different datasets using a little number of parameters such as DOSER \cite{zwicklbauer_doser_2016}, AGISTICS \cite{usbeck_agdistis_2014}. In this work, we follow the last two systems and use a collective graph-based approach but with almost no parameters. 

The graph-based methods construct a graph from the candidate entities for each mention in a document and apply a centrality method  in order to get the best entity for each mention.


Some studies have shown that the entity popularity, which considers that the best entity is the most frequent entity for a given mention, is a strong baseline for entity linking\cite{hachey_evaluating_2013,cheng_relational_2013}. It is also very fast method therefore in contrast to other similar works, instead of replacing the popularity method with a collective graph-based approach, we propose to improve the popularity by using a collective graph-based method. 

Hence, we assume that the best entity exists in the top x results given by the popularity method, and we take only these x results to use them for constructing the disambiguation graph which minimises the noise in the graph as many unsuitable nodes and edges will be avoided and the size of the graph will be reduced,  which leads to  more accurate and  faster system, such system can be efficiently applied to Big Data.

In this work,  we exploit the anchor texts in Wikipedia to construct a mention-entity dictionary in order to be used for candidate selection. We also extract the outgoing linking from each Wikipedia pages which refers to another Wikipedia page to construct entity-entity dictionary in order to be used for constructing the disambiguation graph for each document. The mention-entity dictionary will be used to select the top candidates for each mention, then we construct a graph between these candidates using the links from the entity-entity dictionary, afterthat five centrality measures will be applied on the graph, each one gives a score for each entity, a mention will be linked to the entity which gets the most relevant score among other candidates. We also compare the performance of these five methods on four different datasets in order to choose the best centrality measure overall datasets.

The rest of this paper is organised as follows: Section 2 is devoted for problem definition and formulation, Section 3 gets into the related work, Section 4 describes the proposed system, Section 5 presents the centrality measures which we have applied, Section 5 describes the datasets,  experiments and results and Section 6 concludes the paper with some future perspectives.

\section{Problem Definition and Formulation}

Let $M=<m1, m2, ...., mn>$ be a set of mentions in a document d. Let K be a knowledge base (KB), and $E=<e1, e2, ..,en>$ be a set of entities from K which represent the mentions: m1 refers to e1, m2 to e2 and so forth.  Our problem is to find the best E for a given M specifically the best entity ei for each mention mi, therefore we aim to find a mapping between each mention and entity given the other mentions and  K.

$$ei= argmax_{e \in K} {P(e|m_{i},K,M)}$$

We approximate the solution by decomposing the problem: we first limit the number of entities by candidate selection phase which selects a few number instead of all K, then we compute $p(e|m_{i})$ which is the popularity method to select the most relevant x candidate entities to be used for  constructing a graph  in order to pick up the most relevant entity. Thus, the calculation of  $ {P(e|m_{i},K,M)}$ is approximated by selecting the top x candidates for each mention, then reranking them after constructing a disambiguation graph of all candidates in M.

\section{Related Work}
NEL consists of two phases, the first one is candidate selection which aims to select the most relevant entities for a mention from a knowledge base. Most research have exploited the link structure  in Wikipedia for constructing a mention-entity dictionary where  each mention may refer to different entities. For example,   In AIDA \cite{hoffart_robust_2011}, authors proposed "Yago Means" which is   derived from Wikipedia,  DOSER \cite{zwicklbauer_doser_2016} and  AGISTICS \cite{usbeck_agdistis_2014}  extracted the rdfs:label attribute of DBpedia which is also derived from Wikipedia.  \cite{chang_comparison_2016} constructed 4 dictionaries  from Wikipedia, then a strategy to  disambiguate the mentions have been proposed  to use  these dictionaries.  Others have used a corpus of web links, \cite{chisholm_entity_2015} showed that using 34 millions web links instead of Wikipedia gives a similar performance but  combining Wikipedia with web links outperforms both of them. 

In this work we exploit the link structure of Wikipedia for building a mention-entity dictionary, we do not go farther for this phase because we are actually more interested in exploring the second phase.

The second phase is the candidate  disambiguation in which one entity will be chosen from the candidate set for each mention. There are two main approaches for this phase: single mention disambiguation and collective disambiguation. 

The first one disambiguates one entity mention at a time without considering the effect of other entity mentions, Support Vector Machine has been firstly used \cite{bunescu_using_2006}, then   a large-scale system for entity disambiguation by incorporating  different  features  has been presented \cite{cucerzan_large-scale_2007}, statistical methods with rich relational analysis has been incorporating \cite{cheng_relational_2013}, and recently, Convolotional neural networks  that compute the similarity between a mention and an entity have been proposed  \cite{francis-landau_capturing_2016}.

  The collective approach disambiguates jointly all entity mentions, it tries to model the interdependencies between the candidate entities for all mentions. This approach reformulates the problem as a global optimisation problem which is NP-hard for which many approximations have been proposed.
\cite{ratinov_local_2011} proposed a local and global methods to address entity linking problem, while the local methods take into account the similarity between the mention and the candidate entity, the global one is based on the entire document and all mentions were disambiguated concurrently in order to produce coherent results.
  \cite{hoffart_robust_2011} built a weighted graph of mentions and candidate entities, and computes a dense subgraph that approximates the best joint mention-entity mapping. 
  A probabilistic method that makes use of graphical models to perform collective disambiguation  \cite{ganea_probabilistic_2016}.
\cite{alhelbawy_graph_2014} used PageRank to rank the candidate entity,  \cite{pershina_personalized_2015} proposed Personalized PageRank to filter out the noise introduced by incorrect candidate entities.
  AGISTICS \cite{usbeck_agdistis_2014} used HITS algorithm to rank the graph nodes, while DOSER \cite{zwicklbauer_doser_2016} found that PageRank performs better on their disambiguation graph.  \cite{brando_reden:_2016} proposed different centrality measures to collectively disambiguate the authors's names in a French corpus.

Different works have concluded that choosing the most popular candidate entity is a strong baseline and it is difficult to beat \cite{hachey_evaluating_2013,cheng_relational_2013}. Here, we are based on this observation therefore we propose to improve the popularity method by constructing a graph between the top candidate entities which  expected to reduce the noise in the graph and is not  time-consuming as when using all candidate entities. HITS, PageRank, Degree, Betweenness, Closeness are compared to obtain the most robust measure which works well over different datasets.


\section{Proposed System}

In this section, we present how we use Wikipedia for candidate selection and Disambiguation graph construction.
 \subsection{Candidate Selection}
 For selecting the candidate entities for each mention, we construct a mention-entity pairs dictionary from English Wikipedia, we extract the title of each page and parse the page content to extract its outgoing links (anchor texts). We consider that entities are the Wikipedia page titles, each title  is a mention-entity pair, both the mention and the entity are the title itself\footnote{A title is an entity, but also it could be used to refer to itself therefore it is  also a possible mention.}. Each outgoing link is composed of an anchor text and a link to another Wikipedia page, the anchor text is the mention and the link is the entity. Table 1 shows an example for the  top 3 candidate entities for the mention "sun"  ordered by their occurrences in Wikipedia.  Table 2 shows some statistics of  Wikipedia version which is used to construct the mention-entity dictionary.

 \begin{table}[th!]
\centering
\begin{tabular}{|c|c|c|}
\hline
Mention & Page Title& Count\\
\hline
 sun    &  The\_sun\_(United\_kingdom)                                                                      &     4692\\
 sun    &  Sun\_Microsystems                                                                    &      230\\
 sun    &  Planet\_in\_astrology                                                                     &   59 \\
\hline
\end{tabular}
\caption{A part of mention-entity dictionary where mention is Sun. The full URL of the Wiki page is https://en.wikipedia.org/wiki/ followed by the  Page Title. }
\end{table}

\begin{table}[h!]
\centering
\begin{tabular}{|c|c|}
\hline

Number of Pages & 5,206,974\\
Number o Pages having links & 5,191,234\\
 Number o Mentions & 74,753,045\\
  Number o Distinct mentions  &10,584,594\\
\hline

\end{tabular}
\caption{Statistics on English Wikipedia.}
\end{table}
 \subsection{Disambiguation Graph Construction}
 The graph-based method constructs a  disambiguation graph  where each node represents a candidate entity and each edge represents a link between two entities. Therefore, we need  to find the links between the candidate entities. For this purpose, we also leverages Wikipedia for extracting entity-entity relations dictionary. We parse each Wikipedia page to extract all outgoing links, and for each link we add an entry to the dictionary where the first entity is the actual Wikipedia page and the second one is the target page.  Table 3 shows the most frequent outgoing links from Sun\_Microsystems page.
  \begin{table}[th!]
\centering
\begin{tabular}{|c|c|}
\hline
Actual Page & Target Page \\
\hline
 Sun\_Microsystems  & Oracle\_Corporation                                                                            \\
 Sun\_Microsystems  & StarOffice                                                           \\
 Sun\_Microsystems  & Java\_(programming\_language)                                                       \\
\hline
\end{tabular}
\caption{A part of entity-entity dictionary where the underlying page is Sun\_Microsystems. }
\end{table}

Figure 1 shows the disambiguation graph for the sentence "Java was created by Sun." where there are two mentions: Java and Sun. For each mention we take the top most popular candidates using mention-entity dictionary, then for each candidate we look up in the entity-entity dictionary to find if there is a link to another  entity in the graph, therefore a bidirectional link is made from Sun\_Microsystems to Java\_(programming\_language).  

 \begin{figure}
 \centering
\includegraphics[width=8cm, height=2.5cm]{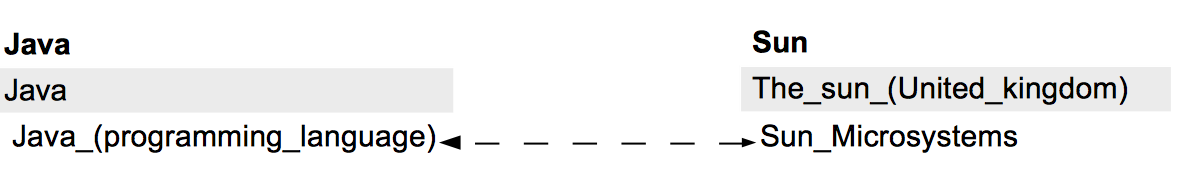}
\caption{Disambiguation graph example.}
\label{Tux}
\end{figure}

 \subsection{Proposed Methods}
 
Popularity-Based  approach treats each mention independently, we compute the popularity of an entity given a mention. The popularity in this regard is the conditional probability which is based on the number of the occurrences of the entity-mention pair in Wikipedia normalized by the occurrences of all entities. Then we select the most popular entity for each mention. This approach does not take into account the context. To surpass this limitation we construct a disambiguation graph to improve the accuracy of this method. 

Constructing a disambiguation graph is very cost for each  document, so we propose to use the popularity to limit the number of  candidate entities set for a mention to only 3. The reason behind that is that in most cases the correct entity for a mention appears in the first three candidates and  using more number of candidates is not useful because the added entities and edges will increase the ambiguity in the graph and negatively affect the results (it should note that some datasets are not affected by increasing the number of candidates such AIDA-CONLL).
After constructing the disambiguation graph, we apply five centrality measures to compute a score for each candidate entity. Later on, we compute the average accuracy for each measure over 4 datasets to get the most reliable measure.

\section{Centrality Measures}
  Once the directed graph is constructed, centrality measures are computed to assign a score to each node. We  use five centrality measures:
  \subsection{Degree}
	  The degree centrality for a node e is the fraction of nodes it is connected to. 
  
	  $$degree(e)=\frac{number\ of\ incoming\ links}{n-1}$$
  
 	The degree centrality values are normalized by dividing by the maximum possible degree in a  graph n-1 where n is the total number of nodes. Degree is the first and  the simplest  centrality measure.
  \subsection{HITS}
  Hyperlink-Induced Topic Search (HITS)  \cite{kleinberg_authoritative_1999} computes two numbers for a node: Authorities estimates the node value based on the incoming links and  Hubs estimates the node value based on outgoing links. It supposes that good authorities are pointed to by good hubs and good hubs point to good authorities. HITS starts with initial two values for each node, then it  successively refines them. Let $x_{i}$, $y_{i}$ is the authority and hub scores for a node i, E the set of directed edges in the graph, $e_{ij}$ represents the directed edge from node i to node j.
  $$x_{i}^{(k)}=\sum_{j:e_{ji}\in E} {y_{j}^{(k-1)}} \  \ , \ \  y_{i}^{(k)}=\sum_{j:e_{ij}\in E} {x_{j}^{(k)}} $$
  $$ \  \  for \ k \ in \ 1,2,...
  $$
  \subsection{PageRank}
  PageRank \cite{page_pagerank_1998} computes a ranking of the nodes in the graph  based on the structure of the incoming links, the underlying assumption is that more important nodes are likely to receive more links from other nodes. It starts by assigning an initial score for each node then it successively refines them .
$$pagerank(i)^{(k)}=\sum_{j:e_{ji}\in E} {\frac {pagerank(j)^{(k-1)}} {|j|}}$$ $$ \  \  \  for\ k \ in \ 1,2,...
  $$
  $|j|$ represents the number of outgoing links from j.

  \subsection{Betweenness}
     Betweenness centrality measure \cite{brandes_faster_2001} is based on shortest paths, it is  the number of these shortest paths that pass through the node. [Freeman (1977) REF] It is given for a node i by the expression:

$$betweenness(i)=\sum_{s\neq i\neq t \in E}{\frac  {Q_{ts}(i)} {Q_{ts}}}$$
 
 where $Q_{ts}$ is total number of shortest paths from node s to node t and $Q_{ts}(i)$  is the number of those paths that pass through i.

      \subsection{Closeness}
       The more central a node is, the closer it is to all other nodes.  Thus, Closeness was defined  as the reciprocal of the fareness \cite{freeman_centrality_1978}:
                
          $$closeness(i)=\frac{1}{\sum_{j \in E} {d(j,i)}}$$

       where d(j,i) is the distance between node  i and j.

\section{Experiments and Evaluation}
The proposed system is implemented in Python, NetworkX library is used for  graph construction and centrality measures. Four datasets are used to evaluate it and get the most efficient centrality measure. In this section, we describe the used datasets and the experiments which we have done and the results \footnote{The source code will be soon available on GitHub.}.
\subsection{Datasets}
For evaluating our systems, we have chosen 4 well-known and publicly available datasets:
\begin{enumerate}
\item ACE2004: This dataset is a subset of ACE2004 coreference documents which contains 253 mentions in  56 news articles \cite{ratinov_local_2011}.
\item AIDA/CONLL-TestB: This dataset contains 231 news articles, it is derived from CONLL-2003 shared task and annotated by \cite{hoffart_robust_2011}.
\item MSNBC: This dataset contains  20 news articles, it comprises a wide range of entities  \cite{cucerzan_large-scale_2007}.
\item Microposts-2014-Test: This tweet dataset contains 687 tweets \cite{usbeck_gerbil_2015}.
\end{enumerate}

\begin{table*}[!h]
\centering
\begin{tabular}{|c|c|c|c|c|}
    \hline 
  & AIDA/CONLL-TestB& ACE2004& MSNBC& Microposts-2014\\
   \hline 
    Number of documents 		         & 231   & 36  & 20   & 687 \\
    Number of mentions per document&19.1     &  8.1    &36.2   & 2 \\
    Number of mentions                       & 4420& 290 & 723 & 1405\\
    mentions with candidate        & 4155 & 216 & 460 & 1091\\

        \hline  
\end{tabular}
\caption{Datasets statistics. }
\end{table*}

All annotations in these datasets refer to Wikipedia or DBpedia. Both of  them contain the same entities whose URLS can be easily converted from one to another. Some statistics about these datasets are shown in Table 4. It should note that we removed all the mentions which their annotated entities does not exist in the Wikipedia version which we use.
\subsection{Experiments and Results}
For evaluating our proposed method, we construct the disambiguation graph using the first top 3 candidate entities for each mention, then we use each centrality measure to compute the importance of each entity. For each mention we choose the entity which has the highest score among all 3 candidates. The following Table 5 shows the results in terms of micro-accuracy (percentage of mentions linked correctly) and f-score. 
We have evaluated the two stages: Overall performance which shows the total performance of the system over the two stages. The disambiguation performance in which we remove all the mentions which do not have the correct entity in their candidate set, because in this case the disambiguation algorithm will certainly not be capable to get the correct answer therefore we should not penalise this algorithms because of the weakness of the candidate selection phase. 

For AIDA/CONLL-TestB, the popularity method achieves 65.78\% in terms of overall accuracy, all centrality measures improve it, the best one is the Degree centrality (+14\%), the worst one is the Closeness (+6\%). We remark that the centrality measures affect this dataset more than others datasets, one possible reason is that the entities in this dataset are more connected through wikipedia. The Degree is also the best for disambiguation accuracy, it improves the performance (+14\%). The popularity method achieves 68.82 (+3\%) more than the overall accuracy, which means that the selection phase is fair in this dataset (few mentions do not contain the correct entity in their candidate set).

The popularity for MSNBC achieves 61.19\% (5\% less than AIDA/CONLL-TestB), all centrality measures improve this baseline, the best one is HITS (72.2\%) then Degree (71.27\%), the worst one is Closeness (65.86\%). The disambiguation accuracy for popularity is  78.6\%, 17\% more than the overall one which means that the selection phase is not fair enough for this dataset, more work on this phase should be done to improve the results. HITS is still the best with 90.23\% and Degree (86.92\% ) is 4\% lower  while it was 1\% lower with overall accuracy  which means that some incorrect entity candidates for the  removed mentions may affect positively the results, that leads us to think more about the disambiguation graph and how we can minimise the impact of incorrect candidate entities.  The results reach 90\%, therefore  the graph-base centrality works very well on this dataset  where the average number of mentions per document is 36.2.
 
 For ACE2004, the popularity gives 78.42 \% and all measures improve it. The best one is Degree with 81.33\% and the worst are  Betweenness and Closeness (79.25\%). The disambiguation accuracy for popularity is 87.84\%, 9\% more than overall accuracy which also indicates the importance of selection phase. Degree centrality is also the best 89.19\%. The centrality measures improve the results in this dataset up to 3\% while in the previous two datasets were up to 9\% and 15\%. Therefore, centrality measures does not enough fair for this dataset in which the the average number of mentions per document is 8.1.

Likewise for Microposts-2014-Test, the popularity gives 56.09 \% which is very low comparing with other datasets and all measures improve it a little bit. The best one is Degree with 58.08\% and the worst is  Betweenness  (56.8\%). The disambiguation accuracy for popularity is 69.65\%, 13\% more than overall accuracy which also indicates the importance of selection phase. Degree centrality is also the best (72.6\% )  centrality measures. Therefore, centrality measures does not enough fair for this dataset in which the document is very short and  the average number of mentions per document is 2.

\begin{table}[!h]
\centering
\begin{tabular}{|c|c|c|}
    \hline 
Method&  Overall Accuracy&  Dis. Accuracy\\
   \hline 
     popularity&66.76 &76.23  \\
      hits&  \textbf{72.54} & 83.23  \\
      degree &\textbf{72.55}&  \textbf{86.55}  \\
      pagerank&   70.74 &83.37 \\
      betweenness &  70.24 &81.68\\
      closeness&65.05& 79.55\\
     \hline  
\end{tabular}
\caption{Average Overall  and Disambiguation Accuracy over the 4 dataset for each centrality measure. }
\end{table}

\begin{table*}[!h]
\centering
\resizebox{\textwidth}{!}{%
\begin{tabular}{|l|l||c|c|c|c||c|c|c|c|c|}
    \hline 
   \multirow{2}{*}{Dataset} &  \multirow{2}{*}{Method} &\multicolumn{4} {c||}{OverAll}  & \multicolumn{4} {c|}{Disambiguation}\\
    \cline{3-10}
&  &Accuracy& F Score &Precision& Recall &Accuracy& F Score &Precision& Recall\\
     \hline      
      Aida  Test-B & popularity &  65.78  &   68.94  &  67.57  &  70.36 &  68.82  &   71.1  &  70.35  &  71.87  \\
 & hits &  79.19  &   80.93  &  80.13  &  81.74  &  81.89  &   82.55  &  82.48  &  82.61  \\
 & degree &  \textbf{79.53}  &   81.01  &  80.26  &  81.78 &  \textbf{83.01}  &   83.21  &  83.24  &  83.19  \\
 & pagerank &  73.81  &   76.29  &  75.43  &  77.17  &  77.22  &   78.67  &  78.46  &  78.87  \\
 & betweenness &  73.65  &   74.65  &  73.86  &  75.46  &  78.06  &   78.39  &  78.33  &  78.45  \\
 & closeness &  72.13  &   73.6  &  72.59  &  74.64   &  76.17  &   76.23  &  76.01  &  76.45  \\
            
         \hline
         MSNBC    & popularity &  61.19  &   79.78  &  75.17  &  85.0  &  78.6  &   84.21  &  81.75  &  86.82  \\
 & hits & \textbf{ 72.2}  &   82.66  &  79.43  &  86.15  &  \textbf{90.23}  &   89.23  &  88.55  &  89.92  \\
 & degree &  71.27  &   81.7  &  78.65  &  85.0  &  87.44  &   86.92  &  86.26  &  87.6  \\
 & pagerank &  70.71  &   80.51  &  77.11  &  84.23  &  89.3  &   87.69  &  87.02  &  88.37  \\
 & betweenness &  71.27  &   82.81  &  79.72  &  86.15  &   89.3  &   87.69  &  87.02  &  88.37  \\
 & closeness &  65.86  &   76.24  &  73.14  &  79.62  &  83.72  &   83.97  &  82.71  &  85.27  \\

       \hline  
         	ACE  & popularity &  78.42  &   79.66  &  80.69  &  78.66  &  87.84  &   88.14  &  87.84  &  88.44  \\
 & hits &  80.91  &   81.95  &  83.19  &  80.75 &  88.51  &   88.44  &  88.44  &  88.44  \\
 & degree &  \textbf{81.33}  &   82.38  &  83.62  &  81.17  &  \textbf{89.19}  &   89.12  &  89.12  &  89.12  \\
 & pagerank &  80.91  &   81.95  &  83.19  &  80.75  &  88.51  &   88.44  &  88.44  &  88.44  \\
 & betweenness &  79.25  &   80.25  &  81.47  &  79.08  &  87.84  &   87.76  &  87.76  &  87.76  \\
 & closeness &  79.25  &   80.68  &  81.9  &  79.5  &  87.16  &   87.76  &  87.76  &  87.76  \\

      \hline  
      Tweet  & popularity &  56.09  &   56.62  &  56.42  &  56.81   &  69.65  &   70.04  &  69.74  &  70.35  \\
 & hits &  57.51  &   57.71  &  57.56  &  57.87  &  72.29  &   72.15  &  71.95  &  72.35  \\
 & degree &  \textbf{58.08}  &   58.39  &  58.24  &  58.55  &  \textbf{72.6}  &   72.59  &  72.39  &  72.79  \\
 & pagerank &  57.51  &   57.71  &  57.56  &  57.87  &  72.29  &   72.15  &  71.95  &  72.35  \\
 & betweenness &  56.8  &   57.24  &  57.07  &  57.41   &  70.6  &   70.85  &  70.58  &  71.13  \\
 & closeness &  57.15  &   57.64  &  57.49  &  57.79  &  71.13  &   71.37  &  71.18  &  71.57  \\    
     \hline 
\end{tabular}}
\caption{Results on four datasets using five graph-based methods in addition to the popularity}
\end{table*}

Thus, the centrality measures can improve a strong baseline for entity linking. The Degree centrality measure which is a simple and fast computed one  gives the best results for overall and disambiguation performances.  HITS is very close to Degree in terms of overall performance but it is less useful in terms of overall performance. Table 5 shows the average accuracy for each centrality measure over the four datasets. 

The performance of Degree centrality may be surprising because of the proved performance of HITS and PageRank, but these two algorithms have been designed to work on the Web, they suppose that there is some spam pages and the search engine want to give a lower rank for these pages, in Wikipedia this assumption is not applied, and the pages which have lots of outgoing links and less incoming links may still  important pages, therefore a Degree centrality may be a suitable measure and designing new measures should take in consideration  the difference between  Web and Wikipedia.

The number of mentions per document plays an important role in graph-based method, Constructing one graph for each document may not be the best approach,  constructing a graph for each paragraph  may be a more reliable approach, but  if the paragraph does not have enough number of mentions (as in tweets), the disambiguation graph will not reflect the interactions between the entities, in contrast if it has many number of mentions, the graph will have a lot of noise. Segmenting the document to construct a graph for each segment may  increase the accuracy but also the complexity. The fact that a simple measure as  degree could give a good results make the graph construction  not necessary so that  the segmentation will not decrease the performance of such system.


\section{Conclusion}
We proposed to use  graph-based methods to improve a strong baseline based on popularity for entity linking. We  use the popularity method  as a method to reduce the size of the disambiguation graph.  The experimental results show that the collective methods improve the performance of popularity. Degree centrality, a simple and fast measure gives the best result, the fact that this measure does not need to construct a graph to be computed open the door to construct different disambiguation graphs for each document. Our future work  will focus on constructing different graphs for a document and  more attention will be devoted for selection phase which is critical  to improve the overall performance.
\section{Acknowledgment}
This research is funded by Labex Observatoire de la vie littéraire (OBVIL) and Laboratoire d'Informatique de Paris 6 (LIP6), Pierre and Marie Curie University, UMR 7606, Paris, France.
\bibliography{tacl}
\bibliographystyle{apalike}

\end{document}